\documentclass[letterpaper, 10 pt, conference]{ieeeconf}
\IEEEoverridecommandlockouts
\usepackage{cuted}
\usepackage{siunitx}
\usepackage{cite}
\usepackage{amsmath,amssymb,amsfonts}
\usepackage{algorithmic}
\usepackage{graphicx}
\usepackage{textcomp}
\usepackage{caption}
\usepackage{xcolor}
\usepackage{tabularx}
\usepackage{booktabs}
\usepackage{tikz}
\usepackage{subcaption}
\usepackage{multirow}
\usepackage[most]{tcolorbox}
\usetikzlibrary{arrows.meta}
\usepackage{pgfplots}
\usepackage{float}
\usepgfplotslibrary{groupplots}
\pgfplotsset{compat=1.18}

\makeatletter
\let\NAT@parse\undefined
\makeatother
\usepackage[hidelinks]{hyperref}

\def\BibTeX{{\rm B\kern-.05em{\sc i\kern-.025em b}\kern-.08em
    T\kern-.1667em\lower.7ex\hbox{E}\kern-.125emX}}

\begin{document}

\title{\LARGE \bf
Learning Multi-Phase Robot Loco-Manipulation \\via Phase-Conditioned Policy Optimization
}

\author{
Simone Manni$^{1}$, Annika Wollschläger$^{2}$, Dario Bellicoso$^{2}$, Abel Gawel$^{2}$,\\
Marco Hutter$^{1,2}$, Simon Le Cleac'h$^{2}$, and Jan Brüdigam$^{2}$%
\thanks{$^{1}$Robotic Systems Lab (RSL), ETH Zurich, Zurich, Switzerland.}%
\thanks{$^{2}$Robotics and AI Institute (RAI), Zurich, Switzerland.}%
}

\maketitle

\begin{strip}
    \centering
    \vspace{-1.0cm}
    \includegraphics[width=\textwidth]{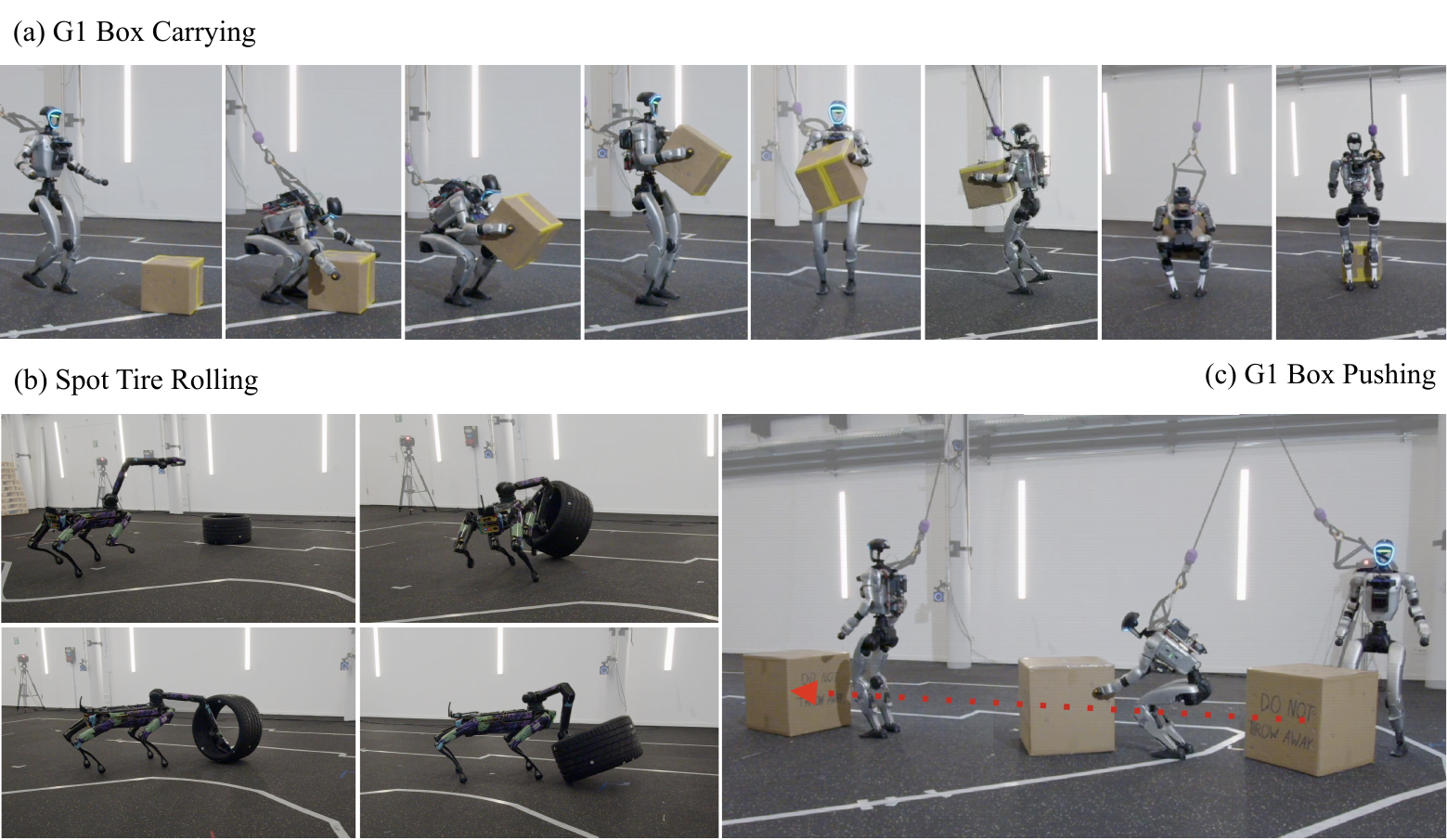}
    \captionof{figure}{\textbf{Multi-Phase Loco-Manipulation.} 
    Our framework synthesizes a single, unified policy per environment capable of executing complex, multi-phase loco-manipulation behaviors, including: 
    (a) a box pick-up and carrying sequence on the Unitree G1 humanoid; 
    (b) a full tire-manipulation sequence (uprighting, rolling, and tilting) on Boston Dynamics Spot; and 
    (c) a dynamic box pushing task with G1.}
    \label{fig:teaser}
\end{strip}

\begin{abstract}
Complex whole-body loco-manipulation requires executing diverse maneuvers, such as crouching, grasping, and locomotion, within a single long-horizon task. While performing simple pick-and-place behaviors is well-established, scaling to complex, long-horizon tasks remains difficult. Currently, this challenge is addressed by relying on human motion data and correspondingly humanoid platforms, which limits platform independence and introduces substantial data processing overhead.
Instead, we extend proximal policy optimization (PPO) to train a single policy for multi-phase tasks across diverse robotic hardware, leveraging prior task structure to shape behavior. Our method adapts PPO using a phase-conditioned critic, phase-gated reward functions, and a multi-phase reset curriculum to guide exploration. To support varied hardware, we employ a decoupled hierarchical control architecture that separates lower-level balancing dynamics from high-level phase execution.
We demonstrate multi-stage loco-manipulation skills on hardware on both the Unitree G1 and a Boston Dynamics Spot equipped with a manipulator arm, and validate our design choices through ablations. Our approach enables the acquisition of complex loco-manipulation behaviors purely through environment interaction while directly incorporating high-level task structure into policy learning.
\end{abstract}

\section{Introduction}

Loco-manipulation is an essential capability for deploying mobile robots in real-world environments, where they must navigate, interact with, and reconfigure their surroundings. Although reinforcement learning (RL) methods have been increasingly employed with success in short-horizon pick-and-place tasks \cite{tang2025deep, jung2026learning}, they usually struggle in long-horizon settings due to the large exploration space and complex contact mechanics \cite{pmlr-v270-zhou25a, pmlr-v305-mao25a}.

Additionally, complex loco-manipulation tasks, such as lifting and carrying a box, mandate coordinating distinct physical modes (walking, crouching, rising) within a unified policy. This poses a major challenge for standard reward tuning, which might cause training to collapse into unwanted local minima, such as learning to kick or drag a box instead of lifting and carrying it. To bridge this gap, recent work increasingly relies on human motion datasets \cite{yang2025omniretarget} and demonstrations, driving the widespread adoption of humanoid platforms in research. While effective, this paradigm requires substantial data processing overhead and results in platform-specific methods that fail to generalize to diverse robotic hardware. More recent approaches have successfully solved humanoid loco-manipulation tasks by learning through autonomous environment interaction, without human guidance \cite{byrd2026adaptmanip, leziart2026learning, zheng2025embracing}. However, they only focus on simple table-to-table object transportation, ignoring the distinct physical transitions, such as low crouching and rising, required to pick up or deposit an object directly from the ground.

Crucially, long-horizon tasks are inherently compositional and naturally break down into a sequence of distinct behavioral stages, each demanding specialized local control. For example, a box-lifting policy can be split into approaching, crouching and lifting phases. Exploiting this inherent multi-phase structure provides a promising route to tackle the exploration challenge without demonstration data. While traditional methods often rely on finite state machines to chain independently trained sub-policies sequentially---a strategy previously applied also to humanoid box-lifting \cite{dao2024sim}---managing transitions across highly dynamic regimes can be problematic. In contrast, consolidating these stages into a single policy natively manages inter-phase transitions and eliminates external orchestration. This strong multi-phase structure and the advantage of having a single policy to handle the entire task highlight the need for an RL framework that can explicitly leverage phase transitions to guide exploration and training.

To this end, we introduce a variant of the popular PPO \cite{schulman2017proximal} algorithm that incorporates the sequential nature of the task. A state machine segments the task into manageable behavior windows and masks MDP terms. The actor is forced to internalize task progression by predicting the current phase, and the critic is conditioned on the active phase to better capture local value changes. We evaluate our framework across two distinct platforms and multiple tasks, including a Spot quadruped maneuvering a tire and a Unitree G1 humanoid carrying a box. For the latter, we also resolve several sim-to-real issues to enable stable, prolonged grasping. Finally, through targeted ablations, we validate the impact of our core algorithmic choices.

Our contributions are as follows.
\begin{itemize}
    \item \textbf{Unified Control Interface:} To ensure compatibility, we first establish a common control interface across both platforms, extending the hierarchical RL framework originally developed for the quadruped to the humanoid.
    \item \textbf{Multi-Phase RL Framework:} We design a multi-phase RL training framework built on top of PPO that conditions value estimation and the MDP on task phases, allowing a single unified policy to coordinate complex multi-stage tasks.
\end{itemize}

\begin{figure*}[t]
    \centering
    \includegraphics[width=1.0\linewidth]{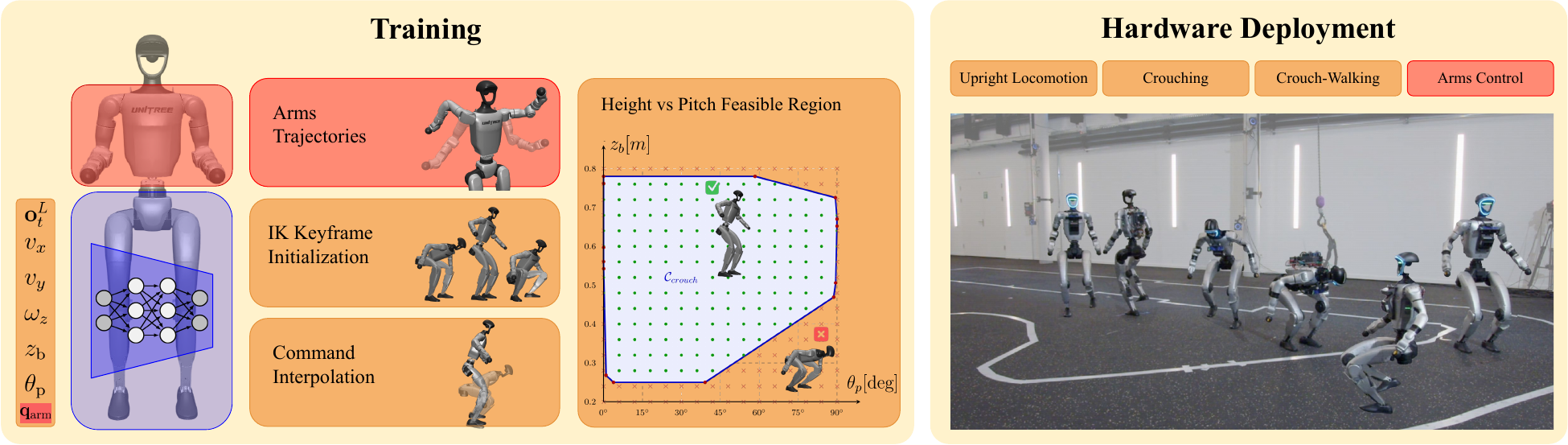}
    \caption{\textbf{Overview of the low-level whole-body policy.} The policy commands leg and waist joints to execute stable locomotion, low crouching, and crouch-walking. Training stability is achieved via inverse kinematics (IK) keyframe initializations that spawn the robot in valid postures, coupled with a feasibility-constrained command space. For standing environments, this region is bounded by the convex hull of admissible IK keyframes, whereas for dynamic environments, the feasible command envelope dynamically contracts as target tracking velocity increases.}
    \label{fig:wholebody_overview}
\end{figure*}

\section{Related Work}
We provide a brief overview of related work on low and high-level policies for robotic manipulation, with a focus on long-horizon tasks.

\subsection{ReLiC for Hierarchical Loco-Manipulation}
Our framework builds on the hierarchical loco-manipulation architecture introduced in ReLiC~\cite{pmlr-v305-zhu25a}, itself grounded in earlier hierarchical RL approaches for quadrupedal loco-manipulation~\cite{jeon2023learning}. ReLiC uses a decoupled hierarchy in which a low-level whole-body controller tracks locomotion and posture commands while a high-level policy specifies task-dependent behaviors. We adopt this architecture as the control interface for our framework. For the Unitree G1, we train a new low-level whole-body controller with the same general design, adapting the interface to the humanoid's kinematics and balancing dynamics. For the Boston Dynamics Spot experiments, we directly use the ReLiC controller. Building on this established hierarchical architecture, our contribution focuses on the high-level learning problem: training a single policy to execute a long-horizon sequence of distinct behaviors.

\subsection{Humanoid Whole-Body Control}
Whole-body control for humanoids has shifted from model-based optimization \cite{1642100, murooka2021humanoid, settimi2016motion} to learning-based policies trained in simulation \cite{gu2026humanoid}. This shift, however, has largely traded one bottleneck for another: rather than hand-designing controllers, the field is now dominated by demonstration-driven, motion-tracking approaches that imitate human motion capture or teleoperated data at increasing scale \cite{zhang2025natural, ji2024exbody2, pmlr-v305-ze25a}. Driven to the extreme, this culminates in foundation-scale efforts such as SONIC \cite{luo2026sonic}, trained on over 100 million frames of motion data. These methods produce highly natural motion but demand substantial data curation and compute. Closer to our approach, ULC \cite{sun2025ulc} trains a single unified policy from scratch using large-scale parallel reinforcement learning, without motion references, to jointly track root velocity, root height, torso rotation, and dual-arm joint positions. We adopt the same demonstration-free, single-policy philosophy, but consider a narrower control objective tailored to our architecture and use cases. Following the decoupled upper- and lower-body control framework introduced in \cite{pmlr-v331-zhang26a}, our policy is trained to track velocity, height, and torso pitch, while observing but not controlling the arms joints, removing the additional torso-rotation and dual-arm tracking objectives considered in ULC. Moreover, while ULC addresses the complexity of its high-dimensional command space through sequential skill acquisition and an adaptive curriculum, we instead constrain commands to feasible regions and use keyframe-based exploration to facilitate learning.

\subsection{Reinforcement Learning for Long-Horizon and Multi-Stage Tasks}

The decomposition of complex, long-horizon tasks builds upon the foundational principles of hierarchical reinforcement learning and the options framework introduced by Sutton and Barto \cite{sutton1999between}. This paradigm relies on learning multiple sub-policies alongside an orchestrator that determines when to switch policy, enabling the agents to operate across distinct levels of abstraction and temporal scales. Subsequent extensions have explored designing the orchestrator as a finite state machine to chain sub-policies for deployment \cite{dao2024sim, konidaris2009skill, chen2023sequential}. However, while traditional skill-chaining methods have succeeded primarily in quasi-static tabletop manipulation, applying them to dynamic loco-manipulation breaks down: defining stable initial and terminal states for sub-policies becomes difficult due to continuous momentum and shifting contacts, which in turn makes both sub-policy training and inter-policy transitions brittle. Hence, they typically require precise state observability at deployment or specialized training regimens \cite{lee2021adversarial, chen2024scar, lee2019composing} to guarantee reliable policy transitions. To handle transitions autonomously, alternative approaches combine sub-policies at training time via distillation \cite{harish2024reinforcement, rudin2025parkour}, although this introduces additional setup overhead. Similarly, our method uses the same idea of splitting a long task into multiple phases but directly optimizes a single policy over them. 

Our formulation also connects to the idea of reward machines \cite{icarte2018using} and their applications in robotic manipulation \cite{baert2024reward}, where individual policies are trained with sub-task-specific rewards. Similarly, our approach allows for phase-dependent rewards, but rather than training separate policies, it leverages this task-splitting intuition to train a single unified policy, utilizing a phase-conditioned value network to correctly capture phase-specific behavior.

\section{Method}
We introduce our phase-conditioned policy optimization method. We first formulate the reinforcement learning (RL) problem, then provide details on the low-level controllers, and finally describe the high-level phased policy learning. 

\begin{figure*}[t]
    \centering
    \includegraphics[width=\linewidth]{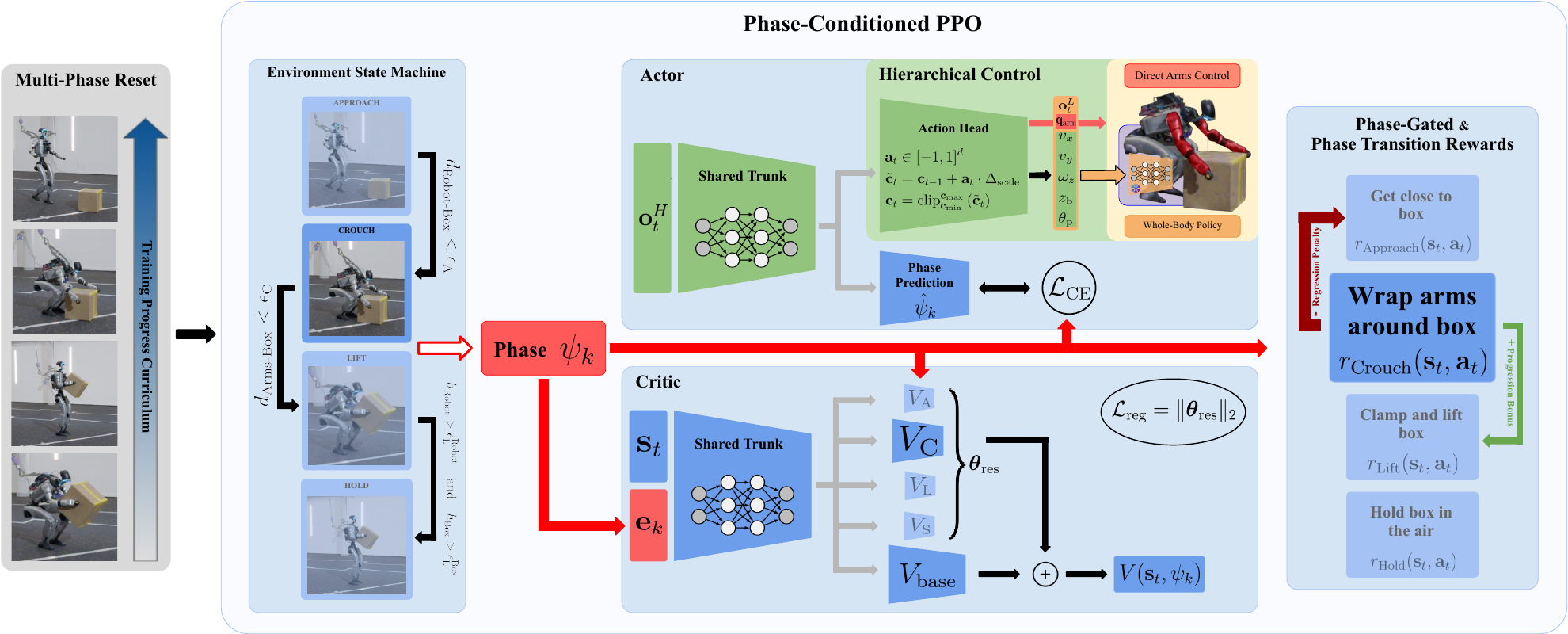}
    \caption{\textbf{Overview of the phase-conditioned PPO framework.} 
    An environment state machine segments the task into phases ($\psi_k$) via local success conditions (e.g., distance $d$ and phase-specific thresholds $\epsilon$).
    To tackle exploration bottlenecks, a multi-phase curriculum initially resets episodes into later stages before progressively shifting back to the initial phase as training advances. 
    The actor outputs delta actions updating command targets $\mathbf{c}_t$, while an auxiliary head enforces phase internalization via cross-entropy loss ($\mathcal{L}_{\text{CE}}$) against the true environment phase. 
    The critic conditions value estimation on state $\mathbf{s}_t$ and phase one-hot encoding $\mathbf{e}_k$, combining a base value $V_{\text{base}}$ with phase-specific residual heads. These are constrained by $\mathcal{L}_{\text{reg}}$ to keep the main learning burden on the base head.
    Finally, task rewards are dynamically gated based on the active phase $\psi_k$ and combined with rewards for forward phase transitions and penalties for backward regression.}
    \label{fig:phasedPPO_overview}
\end{figure*}

\subsection{Reinforcement Learning Problem}

We formulate the long-horizon loco-manipulation task as a partially observable Markov decision process (POMDP) defined by the tuple $\mathcal{M} = (\mathcal{S}, \mathcal{O}, \mathcal{A}, \mathcal{P}, \mathcal{R}, \gamma)$, where $\mathcal{S}$ denotes the state space, $\mathcal{O}$ represents the observation space, $\mathcal{A}$ is the action space, $\mathcal{P}: \mathcal{S} \times \mathcal{A} \times \mathcal{S} \rightarrow [0, 1]$ models the stochastic transition probability density, $\mathcal{R}: \mathcal{S} \times \mathcal{A} \rightarrow \mathbb{R}$ is the reward function, and $\gamma \in [0, 1)$ is the discount factor. At each discrete time step $t$, the agent perceives an observation $\boldsymbol{o}_t \in \mathcal{O}$ derived from the underlying environment state $\boldsymbol{s}_t \in \mathcal{S}$, and samples an action $\boldsymbol{a}_t \in \mathcal{A}$ according to its policy $\pi(\boldsymbol{a}_t \mid \boldsymbol{o}_t)$, generating a trajectory $\tau = (\boldsymbol{s}_0, \boldsymbol{a}_0, \boldsymbol{s}_1, \boldsymbol{a}_1, \dots)$. The objective is to maximize the expected discounted return:
\begin{equation}
    J(\pi) = \mathbb{E}_{\tau \sim \pi} \left[ \sum_{t=0}^{\infty} \gamma^t \mathcal{R}(\boldsymbol{s}_t, \boldsymbol{a}_t) \right].
\end{equation}

To mitigate the intractable exploration space of end-to-end high-dimensional motor control, we adopt a two-level hierarchical RL paradigm. We decompose the controller into a low-level whole-body controller $\pi_{\text{L}}$ and a high-level task planning policy $\pi_{\textit{H}}$.

\subsection{Low-Level Whole-Body Policy}
The low-level whole-body policy $\pi_L$ is designed to track high-level commands, such as velocity and orientation, while observing but not directly controlling the arm joints, which are instead managed by the high-level planner. Adopting an architecture identical to the Spot implementation in ReLiC (omitting the controllable leg), our objective is to establish an efficient, easily trainable policy capable of diverse movements such as walking, crouching, and pitching, entirely without relying on human motion data or trajectory tracking. 

Specifically, $\pi_L$ regulates the $15$ leg and waist degrees of freedom (DoF) to actively track a $5$D command vector $\mathbf{c}_t = [v_x, v_y, \omega_z, z_{\text{base}}, \theta_{\text{pitch}}]^T$, representing planar velocities, pelvis height, and torso pitch, respectively. The policy is implemented as a simple $3$-layer MLP with shapes $[512, 256, 128]$. For comprehensive details regarding arm joint control during training, the observation space (summarized in Table~\ref{tab:observation_space}), and general implementation, we refer the reader to ReLiC \cite{pmlr-v305-zhu25a}. 

For the humanoid platform, key implementation details include:
\begin{itemize}
    \item \textbf{IK-Guided Initializations:} We leverage inverse kinematics (IK)-generated keyframes to spawn the robot in valid postures across diverse base heights and torso pitches. This helps overcome low-crouching exploration traps. 
    \item \textbf{Feasibility-Constrained Command Space:} Pelvis height and torso pitch commands are bounded within valid regions, defined by the convex hull of valid IK keyframes for standing, and scaled by velocity for moving environments.
\end{itemize}

\begin{table}[t]
\centering
\caption{Observation Space Specifications for humanoid low-level whole-body policy}
\label{tab:observation_space}
\small
\begin{tabularx}{\columnwidth}{@{}lXc@{}}
\toprule
\textbf{Group} & \textbf{Description} & \textbf{Dim.} \\
\midrule
Pelvis IMU & Angular velocity \& projected gravity & 6 \\
Torso IMU & Angular velocity \& projected gravity & 6 \\
\midrule
Commands & Planar linear \& angular velocity targets & 3 \\
         & Base height \& pitch targets & 2 \\
         & Arm target joint configurations & 14 \\
\midrule
Joints & Joint positions \& velocities & 58 \\
Memory & Network action at $t-1$ & 15 \\
\midrule
\textbf{Total Dim.} & & $\mathbf{104}$ \\
\bottomrule
\end{tabularx}
\end{table}

To interface $\pi_L$ with the high-level policy, we parameterize the high-level action $\mathbf{a}_t$ as residual delta targets rather than absolute setpoints. Commands $\mathbf{c}_t$ are updated recursively as:
\begin{equation}
    \mathbf{c}_t = \text{clip}\left(\mathbf{c}_{t-1} + \mathbf{a}_t \cdot \Delta_{\text{scale}}, \, \mathbf{c}_{\text{min}}, \, \mathbf{c}_{\text{max}}\right),
\end{equation}
where $\Delta_{\text{scale}}$ scales the step size and $[\mathbf{c}_{\text{min}}, \mathbf{c}_{\text{max}}]$ bound the robot's physical limits. This delta formulation enforces command smoothness by design, eliminating the need for heavy action-rate penalties.

\begin{table*}[t]
\caption{Task-specific behavioral phases across the evaluated robotic platforms and environments.}
\label{tab:task_phases}
\footnotesize
\setlength{\tabcolsep}{2pt}
\begin{tabularx}{\textwidth}{@{\extracolsep{\fill}}llcl@{\extracolsep{0pt}}}
\toprule
\textbf{Task Name} & \textbf{Platform} & \textbf{Phases ($N$)} & \textbf{Behavioral Phases ($\psi_k$)} \\ 
\midrule
\texttt{spot\_tire\_rolling} & Spot & 4 & APPROACH $\rightarrow$ UPRIGHT $\rightarrow$ ROLL $\rightarrow$ TILT-DOWN \\ 
\midrule
\texttt{g1\_push\_box} & G1 & 3 & APPROACH $\rightarrow$ PUSH $\rightarrow$ RISE \\ 
\texttt{g1\_push\_two\_boxes} & G1 & 5 & APPROACH-1 $\rightarrow$ PUSH-1 $\rightarrow$ APPROACH-2 $\rightarrow$ PUSH-2 $\rightarrow$ RISE \\ 
\midrule
\texttt{g1\_lift\_box} & G1 & 5 & APPROACH $\rightarrow$ CROUCH $\rightarrow$ CLAMP $\rightarrow$ LIFT $\rightarrow$ HOLD \\ 
\texttt{g1\_lift\_carry\_box} & G1 & 7 & APPROACH $\rightarrow$ CROUCH $\rightarrow$ CLAMP $\rightarrow$ LIFT $\rightarrow$ CARRY $\rightarrow$ OFFLOAD $\rightarrow$ RISE \\ 
\bottomrule
\end{tabularx}
\end{table*}

\subsection{Phase-Conditioned Policy Optimization}
Now that both the Spot and G1 platforms share a uniform control architecture, we establish a shared methodology on top of that. We observe that complex loco-manipulation tasks (such as lifting and carrying a box) feature clearly defined phases dictated by local success conditions, such as the box clearing the ground, which trigger transitions to subsequent phases. Our approach leverages these structural transitions during training to effectively guide the agent. To implement this, we extend the proximal policy optimization (PPO) algorithm---chosen for its ease of hyperparameter tuning and widespread adoption in the robotics community---with the following core elements: 

\begin{itemize}
    \item \textbf{Environment State-Machine}: Splits long-horizon tasks into specialized behavioral windows based on specific conditions, updating a discrete phase variable $\psi_k$. Crucially, each phase represents a set of states sharing a common property (e.g., all states where the robot is further from the box than a given threshold) rather than being governed by a fixed, time-regulated schedule, with every state belonging to exactly one phase. Furthermore, the local success conditions required for phase transitions during training act as structural guidance to enforce correct behavioral sequencing (e.g., ensuring the box is fully lifted above a threshold before walking). Privileged state information available only in simulation (such as contact metrics) can be used to define these phases and their local success conditions. To prevent oscillation near boundaries, we introduce hysteresis into the transitions: the conditions for regression are intentionally relaxed, allowing the agent to dwell longer in the current phase without premature fallback. While this introduces minor state non-uniqueness across phases, it helps the agent better explore the phase. At test time, phases are no longer used, and the actor operates entirely without observing them or being constrained by them, executing the sequence autonomously.
    
    \item \textbf{Phase-Gated Rewards and Progression Incentives}: Reward terms dynamically activate or deactivate based on the current phase variable to shape phase-specific behaviors. For example, during the approaching phase, the reward prioritizes reducing distance to the box, whereas the subsequent crouching phase gates off approach terms and instead rewards minimizing base height while maintaining stability. Furthermore, we provide sparse positive shaping for forward phase transitions while heavily penalizing backward regression, preventing the agent from oscillating back and forth across phase boundaries simply to farm rewards.

    \item \textbf{Auxiliary Phase-Prediction Head}: Integrates a secondary classification head into the actor, sharing a common feature trunk with the action generation head. Because the actor does not observe the active phase, this auxiliary head is optimized via cross-entropy loss against the true environment label $\psi_k$ to steer the shared latent representation toward encoding task-relevant structure, following the general principle of auxiliary tasks for representation learning in RL \cite{jaderberg2016reinforcement}.

    \item \textbf{Phase-Conditioned Residual Value Network}: The critic processes features through a shared trunk into a global base head alongside a parallel stack of phase-specific residual heads, selecting the active head based on the current phase. Motivated by the difficulty of representing value functions over long horizons \cite{lehnert2018value}, the global head captures overall task progression while the phase-specific heads focus on local nuances from phase-specific reward terms. An $L_2$ norm penalty is applied to their weights to keep the primary learning burden on the base head.

    \item \textbf{Multi-Phase Initialization}: Following DeepMimic \cite{peng2018deepmimic}, we recognize the importance of overcoming exploration bottlenecks by resetting the environment directly into randomized task phases, which enables concurrent optimization across the entire behavioral sequence. Similar to their observation that learning complex motions sequentially creates severe exploration traps, such as failing to learn a jump because poor landings yield high penalties before the jump itself can be mastered, we initialize episodes by sampling across task phases (e.g., spawning the robot close to the tire with the tire already upright). For highly complex maneuvers like box-carrying, we use a curated set of keyframes (e.g., holding the box mid-air), which also helps to bypass severe exploration traps to reliably learn force-closure gripping and holding behaviors. A curriculum progressively scales down these mid-episode resets as training advances, eventually returning all initializations to the first phase.

\end{itemize}

\section{Experiments}
We conduct a number of hardware experiments on multiple robot platforms as well as ablations on our design choices to validate our proposed approach.

\subsection{Hardware Deployments}

We evaluate our framework across distinct tasks and two robotic platforms. On the Boston Dynamics Spot quadruped, the task involves maneuvering a tire from the ground to a target location (\texttt{spot\_tire\_rolling}). On the Unitree G1 humanoid, tasks range from single and dual box pushing (\texttt{g1\_push\_box}, \texttt{g1\_push\_two\_boxes}) to more demanding long-horizon bimanual lifting (\texttt{g1\_lift\_box}) and transport (\texttt{g1\_lift\_carry\_box}). This task is particularly challenging, as it requires the robot to continually adjust its height by transitioning between upright and crouching postures, pushing both our algorithmic framework and the physical hardware to their limits. Beyond this multi-stage kinematic complexity, the task involves continuous contact along the robot's entire forearms rather than just the end-effectors, introducing a heavily under-constrained, contact-rich regime that places extreme demands on both software and physical hardware. Table~\ref{tab:task_phases} provides an overview of the evaluated tasks, detailing the behavioral phases and phase counts ($N$).

All policies are trained in the GPU-accelerated \texttt{mjlab} simulation environment \cite{zakka2026mjlablightweightframeworkgpuaccelerated}. Humanoid policies are exported in ONNX format with \texttt{exploy}\cite{exploy2026} and deployed to physical hardware via a custom ROS2 framework, whereas Boston Dynamics Spot utilizes a separate custom control framework. Both setups use an external motion capture system to track robot and object states, and we randomize object mass, size, and friction to bridge the sim-to-real gap.

\subsubsection{Motion-Capture-Free Execution}

\begin{figure}[htbp]
    \centering
    \includegraphics[width=\linewidth]{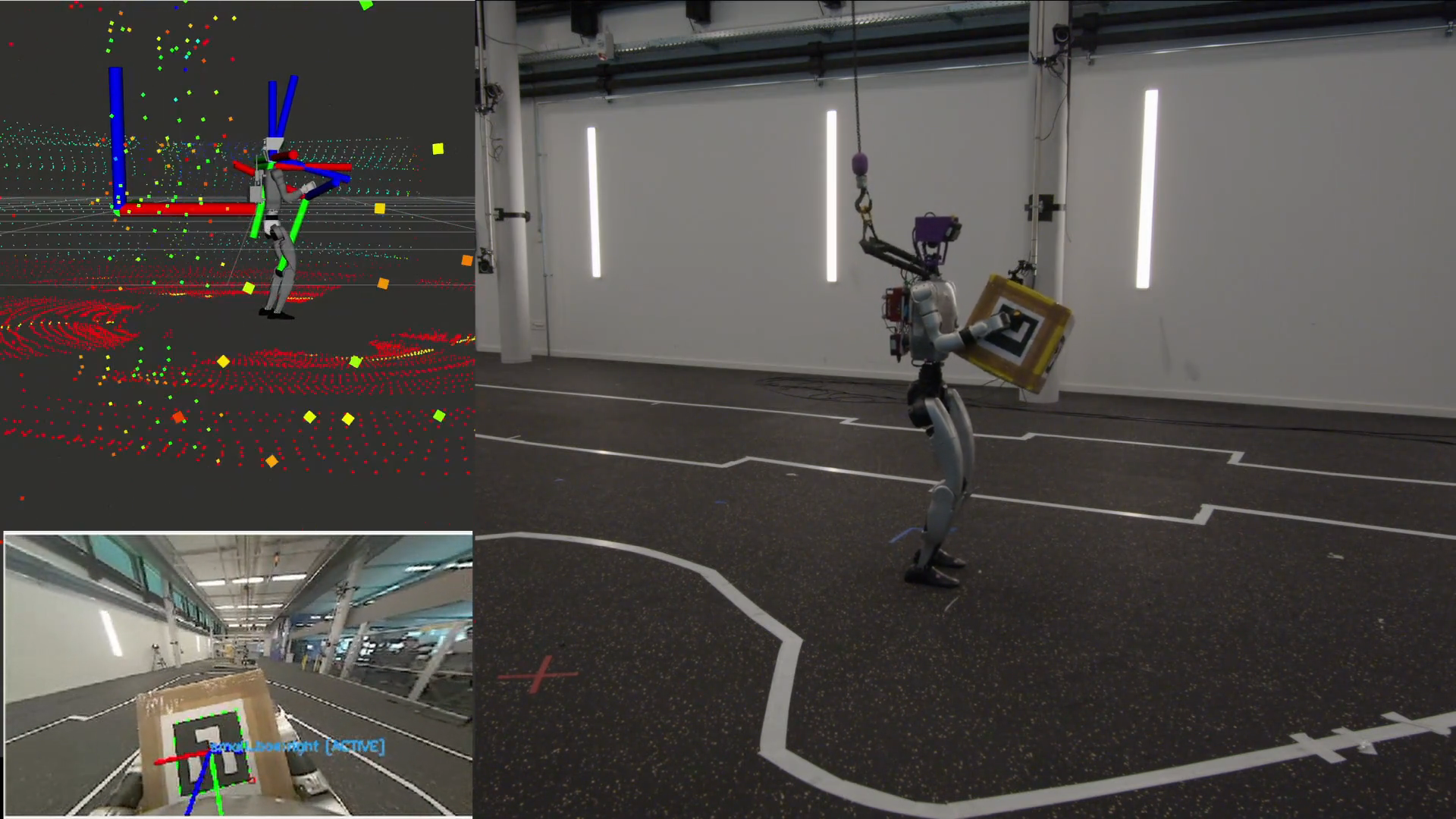}
    \caption{Hardware deployment of the \texttt{g1\_lift\_box} task using onboard state estimation and ArUco marker detection. The right panel shows the Unitree G1 humanoid executing the lift, while the insets display the state estimation visualization (top-left) and the onboard first-person camera view tracking the ArUco marker (bottom-left).}
    \label{fig:g1_lift_box_aruco}
\end{figure}

Because the low-level controller only needs proprioceptive data and the high-level policy relies exclusively on object states expressed in the robot's local frame, our framework is not strictly bound to external tracking infrastructure. By leveraging onboard state estimation and ArUco \cite{garrido2014automatic} marker detection, we successfully execute both the \texttt{g1\_push\_box} and \texttt{g1\_lift\_box} tasks without a motion-capture system, as illustrated in Figure~\ref{fig:g1_lift_box_aruco}. Notably, although training only incorporates standard observation noise without explicitly accounting for missing data or dropouts from marker detection, the policy exhibits sufficient zero-shot robustness to successfully execute the task.

\subsubsection{Sim-to-Real Transfer for Bimanual Grasping}

Physical deployment of the bimanual lifting and carrying policy presented significant challenges due to the contact-rich dynamics of the task, necessitating several key adaptations.

\textbf{Actuator Delays:} Initial hardware evaluations revealed arm oscillations that compromised grasp stability and caused the box to drop. To resolve this, we introduce randomized actuator delays of up to $5$ physics steps (\SI{25}{\ms} at \SI{200}{\hertz}) during training, allowing the policy to compensate for unmodeled high-frequency dynamics and hardware latencies, smoothing trajectory tracking and stabilizing the grasp.

\textbf{Friction Curriculum:} Due to MuJoCo's soft contact model, continuous micro-slippage is an inherent challenge, particularly during extended holding phases required to complete the task. Because MuJoCo Warp lacks native mechanisms to directly resolve this contact artifact, we leverage a friction curriculum: training begins with unrealistically high sliding friction for metal-box contacts (up to 3.0) to facilitate early policy convergence, and anneals it down to similarly elevated final values (1.5–2.0) as training advances.

\textbf{Torque Limits:} The combination of simulation incompressibility and micro-slippage led the policy to exert an excessive gripping force on hardware, which often led to motor overheating (most notably in the shoulder actuators). Rather than applying a reward penalty, which fails to reliably prevent high-force policies and can hinder exploration, we implement a systematic constraint in both simulation and hardware: clamping the maximum allowable deviation between the commanded and current joint positions. Acting as an implicit torque limit, this offset proved essential for safe, prolonged box-clamping, maintaining steady holds for up to 5 minutes. We found that limiting the offset to $7\,\text{N}\cdot\text{m}$ (computed via position-gain $K_p$) struck a good balance between secure clamping and thermal safety.

\subsection{Ablation Analysis}
To justify our architectural choices, we conduct an ablation analysis across two representative tasks: \texttt{spot\_tire\_rolling} and \texttt{g1\_push\_two\_boxes}. We select \texttt{g1\_push\_two\_boxes} because its multi-phase structure provides a rigorous testbed while remaining straightforward enough to isolate architectural performance from sensitive reward tuning, compared to the complex box-carrying task. By systematically removing each component and tracking success rates throughout training, we isolate the specific contribution of each element in the framework.\\

\subsubsection{Auxiliary Phase Prediction}
\begin{figure*}[!t]
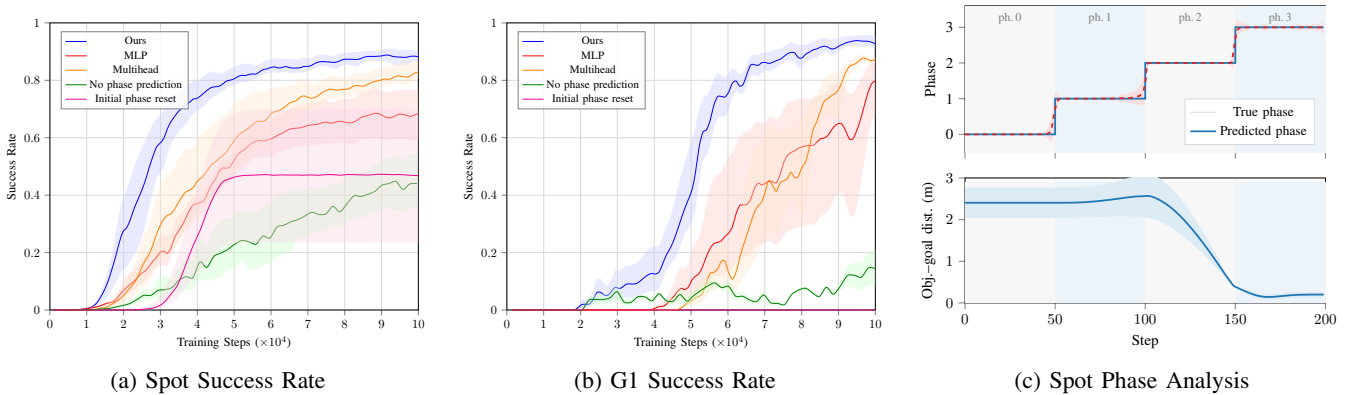

    \centering
    \begin{subfigure}[b]{0.32\textwidth}
        \centering
        \resizebox{\linewidth}{!}{\input{figures/tikz/04_spot_ablation}}
        \caption{Spot Success Rate}
        \label{fig:spot_ablation}
    \end{subfigure}
    \hfill
    \begin{subfigure}[b]{0.32\textwidth}
        \centering
        \resizebox{\linewidth}{!}{\input{figures/tikz/04_g1_ablation}}
        \caption{G1 Success Rate}
        \label{fig:g1_ablation}
    \end{subfigure}
    \hfill
    \begin{subfigure}[b]{0.32\textwidth}
        \centering
        \resizebox{\linewidth}{!}{\input{figures/tikz/04_spot_phase}}
        \caption{Spot Phase Analysis}
        \label{fig:spot_phase}
    \end{subfigure}
    \caption{
    \textbf{Algorithmic Ablation Study on Spot and G1:} Task success rate across training steps on (a) \texttt{spot\_tire\_rolling} and (b) \texttt{g1\_push\_two\_boxes}, comparing our method (\textit{Ours}) against baseline architectures (MLP, Multihead), an unguided variant (\textit{No phase prediction}), and execution without multi-phase episode initialization (\textit{Initial phase reset}). Solid curves and shaded regions denote the mean and variance across 3 random seeds, respectively. \textbf{(c) Spot Phase Analysis:} Evaluation of phase-prediction accuracy across 1,000 successful episodes of \texttt{spot\_tire\_rolling} task. Because individual phase durations vary across rollouts, evaluation episodes are temporally anchored at phase transitions with intermediate timesteps linearly warped to uniform phase durations. The top plot shows that the auxiliary head's online phase predictions tightly track the ground-truth phase steps with minimal variance, while the bottom plot highlights the object-to-goal distance as a proxy for task progression.}
    \label{fig:three_subplots}
\end{figure*}

As shown in Figures~\ref{fig:spot_ablation} and \ref{fig:g1_ablation}, omitting the supervision from the auxiliary phase prediction results in a significantly lower success rate (dropping from $89\% \pm 2\%$ to $44\% \pm 9\%$ on Spot and from $92\% \pm 2\%$ to $14\% \pm 6\%$ on G1) and a reduced convergence speed across both tasks. This performance drop highlights that the auxiliary head plays a crucial role in steering latent representations to cleanly encode discrete behavioral boundaries. For further confirmation, Figure~\ref{fig:spot_phase} displays the phase prediction accuracy over time across 1,000 successful episodes of the \texttt{spot\_tire\_rolling} task. Aside from minor discrepancies at the phase transitions, the plot demonstrates that the head clearly captures the temporal structure of the task. \\

\subsubsection{Multi-Phase Episode Initialization}
To assess the effectiveness of multi-phase initialization, we test the hypothesis that resetting the environment directly into individual phases is essential for acquiring multiple concurrent skills, avoiding policy collapse, and exposing the value function to later transitions. We compare our approach against a baseline trained by initializing the environment exclusively in the first phase. As shown in Figures~\ref{fig:spot_ablation} and \ref{fig:g1_ablation}, while the simpler task achieves success albeit with high variance, the more complex box-pushing task suffers from complete collapse. Specifically, once the policy masters pushing the first box, it encounters completely different rewards and dynamics that it fails to adapt to, because training without multi-phase resets leaves it exposed only to the initial segment of the task. Consequently, multi-phase initialization proves critical for accelerating convergence and maintaining stable learning across extended horizons.\\

\subsubsection{Value Network Topology}
To evaluate whether the architecture combining a standard MLP with phase-specific residuals is necessary, we compare our method against baselines utilizing a simple MLP and fully decoupled heads. While the residual architecture yields a higher success rate and lower variance during training, the quantitative performance difference in standard benchmarks is relatively subtle.

Intuitively, the residual architecture can relieve a normal MLP's parameters from conflicting gradients caused by vastly different phase rewards. Additionally, it also ensures bounded discontinuity at the phase boundaries, which fully decoupled heads fail to do. It is therefore chosen as a balanced middle ground: it allows the network to capture phase-specific local value changes while still internalizing the entire task. Ultimately, the residual head is chosen for these reasons, and it proves successful across all tasks, yielding a higher success rate and qualitatively better policies in the more complex box-carrying task.

\section{Conclusion}
In this work, we introduce an extension of PPO to tackle long-horizon multi-phase loco-manipulation tasks and show its effectiveness on hardware across tasks such as \texttt{spot\_tire\_rolling} and \texttt{g1\_lift\_carry\_box}. 

However, several limitations remain. First, scaling the framework to an increasing number of behavior windows and longer horizons introduces significant representational challenges. Managing a large number of discrete phases can strain policy capacity and complicate value estimation, as the global components of the value network risk overshadowing the fine-grained, local value shifts required for individual phase behavior. Furthermore, the method remains difficult to scale across diverse tasks because it inherently requires manually segmenting the task into explicit phases and introduces additional reward fine-tuning overhead. This is especially pronounced in complex tasks like box-carrying, where changing the box shape requires manually updating reset keyframes, adding friction to scalability. More broadly, loco-manipulation remains constrained by the inherent limitations of simulators, particularly in modeling complex contact dynamics, which compounds the difficulty of transferring policies smoothly to real hardware.

Future work will focus on addressing these algorithmic and scaling bottlenecks. Promising directions include adapting these phase-guided mechanisms to off-policy methods to exploit their superior sample efficiency, as well as extending the framework to collaborative reinforcement learning scenarios for multi-robot coordination. Additionally, future research can explore eliminating manual task segmentation by automatically discovering phase structures from offline datasets, such as those collected via Sample-based Model Predictive Control (SMPC), and leveraging them within offline-to-online reinforcement learning pipelines.

\section*{Acknowledgement}
We thank Maks Sorokin for helpful discussion on policy learning and implementation.

\begin{small}
\IEEEtriggeratref{18}
\bibliography{references}
\bibliographystyle{IEEEtran}
\end{small}

\end{document}